\documentclass[letterpaper, 10 pt, conference]{ieeeconf}  
\pdfoutput=1

\IEEEoverridecommandlockouts                              

\usepackage[latin1]{inputenc}
\usepackage[T1]{fontenc}
\usepackage[english]{babel}
\usepackage{amsmath}
\usepackage{amssymb,amsfonts,textcomp}
\usepackage{array}
\usepackage{hhline}
\usepackage{graphicx}

\usepackage{listings}   
\usepackage{xcolor}
\usepackage{url} 
\usepackage{subfigure}

\definecolor{sh_comment}{rgb}{0.12, 0.38, 0.18 } 
\definecolor{sh_keyword}{rgb}{0.37, 0.08, 0.25}  
\definecolor{sh_string}{rgb}{0.06, 0.10, 0.98} 

\lstdefinestyle{rsg}{
	keywordstyle=\bfseries,
	backgroundcolor=\color{black!10},%
    basicstyle=\small\ttfamily,%
    numbers=left, 
    numberstyle=\tiny, 
    stepnumber=1, 
    numbersep=5pt,%
    stringstyle=\color{sh_string},
    commentstyle=\color{sh_comment}\itshape,
    tabsize=2,
    emph={%
    Box, Cylinder, PointCloud, Node, Group, Transform, GeometricNode, Attribute, child, 
    geometry, transforms, RigidTransform, stamp, value, sizeX, sizeY, sizeZ,
    HomogeneousTransformationMatrix, TimeStamp, cardinality, type, inputStructure,
    inputHook, outputStructure, outputHook, FunctionBlock, PointCloudType,
    sharedPtr, library, root},
    emphstyle={\color{sh_keyword}\bfseries}%
}

\lstdefinestyle{C++}{
	keywordstyle=\bfseries\color{sh_keyword},
	backgroundcolor=\color{black!10},%
    basicstyle=\footnotesize\ttfamily,%
    numbers=left, 
    numberstyle=\tiny, 
    stepnumber=1, 
    numbersep=5pt,%
    stringstyle=\color{sh_string},
    commentstyle=\color{sh_comment}\itshape,
    tabsize=2
}

\begin{document}

\title{\LARGE \bf{
Towards a Domain Specific Language for a \\
Scene Graph based Robotic World Model
}}

\author{Sebastian Blumenthal and Herman Bruyninckx
\thanks{All authors are with the Department of Mechanical Engineering,
		Katholieke Universiteit Leuven, Belgium. Corresponding author:
		{\tt\small sebastian.blumenthal@mech.kuleuven.be}}
}

\maketitle
\thispagestyle{empty}
\pagestyle{empty}


\begin{abstract}

Robot world model representations are a vital part of robotic applications. However,
there is no support for such representations in model-driven engineering tool chains. 
This work proposes a novel Domain Specific Language (DSL) for robotic world
models that are based on the Robot Scene Graph (RSG) approach.
The RSG-DSL can express
(a) application specific scene configurations,
(b) semantic scene structures and
(c) inputs and outputs for the computational entities that are loaded into an instance of a world model.

\end{abstract}


\section{Introduction}
\label{sect::introduction}

Robots interact with the real world by safe navigation and manipulation of the objects of interest. A digital representation of the environment is crucial to fulfill a given task.
Although a world model is a central component of most robotic applications  
a \emph{Domain Specific Language} (DSL) has not been developed yet. 
One reason for this is the lack of a common world model approach. 
The scene graph based world model approach \emph{Robot Scene Graph} (RSG)  \cite{blumenthal13scene} tries to overcome this hurdle.
It acts as a shared resource for a full 3D environment representation in a robotic system. It accounts for dynamic scenes by providing a short-term memory, allows to hierarchically organize scenes, 
supports uncertainty for object poses, has semantic annotations for scene elements and can host computational entities.
While other approaches have a stronger focus on certain world modeling aspects like probabilistic tracking of semantic entities \cite{elfring13semantic} or hierarchical representations for geometric data \cite{wurm11hierarchies}, the RSG emphasizes a holistic view on the world modeling domain.  
This work extends the RSG approach by a \emph{RSG-DSL} to model the structural and computational aspects of the scene graph. It is accompanied by a model to text transformation to generate code for an implementation of the RSG which is a part of the BRICS\_3D C++ open source library \cite{blumenthal13brics_3d}.

A DSL is a formal language that allows to express a certain aspect of a problem domain. It creates an abstraction in order to quickly create new applications and it imposes constraints on a programmer to prevent from programming errors.  
A structured development of a new DSL is organized in four levels of abstractions M0 to M3 \cite{iso05mof}:

\begin{itemize}
\item \textbf{M0}: The M0 level is an instantiation of a DSL model. Typically this results in generated code for a (generic) programming language that can be compiled and executed.
\item \textbf{M1}: The M1 level comprises  models that conform to a certain DSL that is defined on the M2 level.
\item \textbf{M2}: A meta model on the M2 level specifies the DSL in a formal way. This definition has to conform to the meta meta model of M3.
\item \textbf{M3}: The M3 level defines the meta meta model which is a generic model to describe DSLs.
\end{itemize}

The goal of the RSG-DSL for a robotic world model is manifold. This DSL can describe the structural and behavioral parts of a scene that are part of a specific robotic application. It allows to combine the required world model elements at design time. The a priori known structure of a system can include the involved robots with their kinematic structures and their geometries, previously known parts in the environment or the places in the structure where to store online sensor data. Results of the behavioral \emph{function blocks}, which can contain any kind of computation, are stored in the scene graph as well. The selection and the configuration of the function blocks has an important influence on how the world model will appear at runtime. 
For example, the presence of an object recognition function block can enrich the scene graph with task-relevant objects.
The above items can be specified on the code level. 
However, the RSG-DSL reduces the required number of lines of code to encode the scene graph. 
The C++ API assumes a correct order of creation of scene primitives, while the RSG-DSL does not have this restriction.

The proposed DSL allows to express input and output data for the function blocks. This data consists of scene structures to represent parts of the scene graph.
For instance, a segmentation algorithm module consumes a point cloud as input structure and generates a set of new point clouds with associated spatial relations pointing to the center of the segments.

In addition, the RSG-DSL is able to express prior semantic knowledge about a scene. It is possible e.g. to encode a generic version of a table that consists of a table plate and four legs. This can serve as input for a function block that analyzes the perceived scene to recognize that particular structure.

The remainder of the paper is organized as follows: Section \ref{sect::related_work} summaries related work and Section \ref{sect::world_model_primitives} gives a brief introduction to the world model concept. Details of the RSG-DSL are explained in Section \ref{sect::world_model_dsl} and its capabilities are illustrated with examples in Section \ref{sect::examples}. The paper is closed with a conclusion in Section \ref{sect::conclusion}.


\section{Related Work}
\label{sect::related_work}

Recently interest has been risen in robotics to create DSLs for various sub-aspects of robotic systems. 
The Task Frame Formalism DSL \cite{klotzbucher11reusable} has been proposed to describe the control and coordination aspects of robotic software systems. 
A DSL to express geometric relations between rigid bodies \cite{laet13domain} helps to correctly set up spatial relations as constraints will be automatically evaluated on the M1 level. Two DSL variants are discussed: one version is embedded into the Prolog programming language and the second one uses the Eclipse Modeling Framework (EMF) \cite{project13eclipse}. The Prolog approach results in a directly executable code while the EMF variant benefits from the Eclipse tool chain including an editor that supports syntax highlighting and auto-completion. 
The Grasp Domain Definition Language \cite{schneider13towards} is developed in the EMF framework as well. It demonstrates that multiple dedicated robotic languages can be further composed into more complex ones. 

DSL approaches in the 3D computer animation domain have been recently developed for 3D scenes.
The streaming approach for 3D data \cite{haist06adaptive} uses a meta model for scene elements to cope with various 3D scene formants. 
In a similar way the SSIML \cite{lenk12model-driven} approach tries to abstract from the existing 3D formats and APIs method calls. It is meant as a DSL for development of 3D applications. However, in contrast to a robotic world model the complete access to the world state is given. 
To the best of the authors knowledge a DSL for a robotic world model does not exist yet.


\section{World Model Primitives}
\label{sect::world_model_primitives}

		\begin{figure*}[htbp]
			\centering
			\includegraphics[scale=0.88]{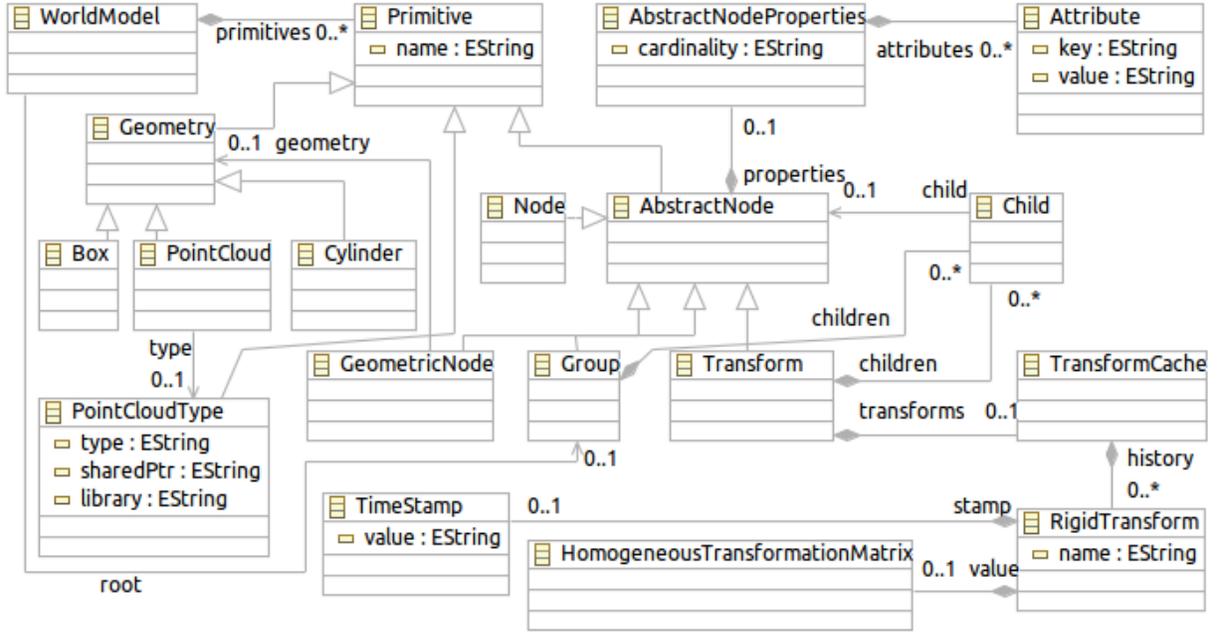}
			\caption{Excerpt from Ecore model of the structural part of the world model. For the sake of readability some elements are not shown: the  	quantities for the geometries and transformation matrix are omitted and the \texttt{Mesh} definition is skipped because it is defined analogous to the \texttt{PointCloud} type.}
			\label{fig::rsg_ecore}
		\end{figure*}

The goal of the world model is to act as a shared resource among multiple involved processes in an application. Such processes could be related to the various robotic domains like planning, perception, control or coordination. To be able to satisfy the needs of the different domains the world model has to offer at least the following set of properties: 		
It appears as shared and possibly distributed resource. It takes the 
dynamic nature and imprecision of sensing of real-world scenes into account, allows for multi-resolution queries and supports annotations with semantic tags.

The scene graph based world model RSG consists of objects and relations among them \cite{blumenthal13scene}. 
These relations are organized in a \emph{Directed Acyclic Graph} (DAG) similar as for approaches used in the computer animation domain. The directed graph allows one to structure a scene in a hierarchical top-down manner. For instance, a table has multiple cups, whereas multiple tables are contained in a room, multiple rooms in a building and so on. Traversals on such a hierarchical structure can stop browsing the graph at a certain granularity to support multi-resolution queries. The graph itself supports four different types of nodes. All node types have in common that each instance has a unique ID, a list of attached attributes for semantic tags and one or more parent nodes. Details of the four node types are given below: 

\begin{itemize}
\item \textbf{Node}: The \texttt{Node} is a generic leaf in the graph. It can be seen as a base class for the other node types.
\item \textbf{GeometricNode}: A \texttt{GeometricNode} is a leaf in the graph that has geometric data like a box, a cylinder,  a point cloud or a triangle mesh. The data is time stamped and \emph{immutable} i.e. once inserted the data connote be altered until deletion to prevent inconsistencies in case multiple processes consume the same geometric data at the same time. 
\item \textbf{Group}: The \texttt{Group} can have child nodes. These parent child relations form the DAG structure.
\item \textbf{Transform}: The \texttt{Transform} is a special \texttt{Group} node that expresses a rigid transform relation between its parents and its children. Each transform node in the scene graph stores the data in a cache with associated time stamps to form a short-term memory.
\end{itemize}

The \texttt{Transforms} are essential to capture the dynamic nature of a scene as changes over time can be tracked by inserting new data into the caches. Moreover, such a short-term memory enables to make predictions on the near future. This requires dedicated algorithms to be executed by the word model as described later. 
In contrast to the \texttt{Transforms}, geometric data is defined to be immutable. Hence, changes on the geometric data structures do not have to be tracked. In case a geometry of a part of a scene does change over time a new \texttt{GeometricNode} would have to be added. The accompanying time stamps still allow to deduce the geometric appearance of a scene at a certain point of time.
All temporal changes in the world model are explicitly represented.

The RSG approach uses a graph structure. Thus, it is possible to store multiple paths formed by the preceding parents to a part of a scene. This case expresses that multiple information to the same entity is available. For example, an object could be detected by two sensors at the same time. Different policies for resolving such situations are possible. Selection of the most promising path like the latest path denoted by the latest time stamps associated with the transforms is one possibility, while choosing a path with the help of the semantic tags is another one. Probabilistic fusion strategies \cite{smith90estimating} are an alternative, given covariance information on the transform data is available. This kind of uncertainty data can be stored in the temporal caches as well. The details of representing uncertainty and fusion strategies are planned as future work.

Besides the structural and temporal aspects, the world model contains \emph{function blocks} to define any kind of computation. A function block consumes and produces scene graph elements. Algorithms for estimating near future states are one example of such a computation. A function block can be loaded as a plugin to the world model and is executed on demand. This allows to move the computation near to the data to improve efficiency of the executed computations. Conceptually the scene graph is a shared resource among all function blocks. Concurrent access to the scene is possible since geometric data is defined to be immutable. \texttt{Transforms} provide a temporal cache such that inserting new data will not affect retrieval of transform data by another function block as long as queries are within the cache limits.

The RSG can be used as a shared resource
in a multi-threaded application. However crossing the
system boundaries of a process or a computer requires
additional communication mechanisms. Many component-based frameworks in robotics including ROS \cite{quigley09ros}, OROCOS \cite{bruyninckx01open}
and YARP \cite{metta06yarp} provide a communication layer for distributed
components. These frameworks are mostly message-oriented
and do not support a shared data structure like a world model well. 
Thus, we allow the RSG to create and maintain local
copies of the scene graph \cite{naef03blue-c}. Subsequent graph updates need to be
encapsulated in the framework specific messages. 
Further details on the RSG world model and its primitives can be found in \cite{blumenthal13scene}.

For a robotic application a set of design decisions has to be made to deploy the shared world model approach. Thus, a DSL for such a world model facilitates the development efforts for a specific application.


\section{A DSL for a Scene Graph based World Model}
\label{sect::world_model_dsl}

\subsection{Choice of modeling framework}

This work uses the Eclipse Modeling Framework (EMF) \cite{project13eclipse} as DSL framework. For two reasons: first, it allows one to make use of the Eclipse tool chain to generate an editor with syntax highlighting. Second, other robotic DSLs that already exist in this framework could be potentially re-used.  
A candidate is the geometric relations DSL \cite{laet13domain}. The integration into the world model DSL is left as future work.

In addition to the proposed DSL, a model to text transformation is provided that generates code to be used in conjunction with the C++ implementation of the RSG which is part of the BRICS\_3D library.  
Hence, this work mainly contributes to the M2 and M0 levels.


\subsection{M2: DSL definition}
\label{sect::dsl_m2}

The RSG-DSL for the scene graph based world model is defined with the Xtext grammar language \cite{xtextproject13xtext}. The corresponding Ecore meta model representation as part of the EMF is completely generated from that Xtext definition. 
The RSG-DSL re-uses an existing DSL for units of measurements that is defined with Xtext as well. This is achieved via \emph{grammar mixins}.

The overall design approach is to find a minimal set of DSL \emph{primitives} and their \emph{relations} that are sufficient to represent the domain of an environment representation that is based on the RSG approach. The core primitives are the different node types and the function blocks. The graph structure allows to relate nodes to each other. The function blocks relate to the graph in the sense that graph structures serves as input and output data structure. 

One central element of the RSG-DSL are the node types as shown in Section \ref{sect::world_model_primitives}. As depicted in Fig. \ref{fig::rsg_ecore} the Ecore model represents the common properties for all node types within the \texttt{AbstractNodeProperties} that can have a list of \texttt{Attributes}. An attribute is a key value pair. The \texttt{Group} and the \texttt{Transform} are the only node types that have \emph{children} by referencing to the \texttt{AbstractNode}. The other two node types \texttt{Node} and \texttt{GeometricNode} are thus leaves in the scene graph.  

The RSG-DSL identifies and references all node types by their names. This seems to be in conflict with the requirement that nodes have unique IDs but the description of a world model on M1 level can be seen as a generic template for a scene of an application \cite{lenk12model-driven}. The constraint of unique IDs has to hold on the M0 instance level and can be considered as an implementation detail. The world model implementation has facilities to provide and maintain unique IDs.
 
Each geometric data that can be contained in a \texttt{GeometricNode} has its dedicated representation within the RSG-DSL. Special attention has to be paid to the \texttt{PointCloud} and \texttt{Mesh} types as legacy data types shall be supported on M0 level. The \texttt{PointCloudType} collects all necessary information to be able to generate code for any point cloud representation used in an application. The \texttt{Mesh} representation follows analogously. On the M0 level this variability is mapped to a template based class.  

The temporal cache for the \texttt{Transform} node is modeled by the \texttt{TransformCache}. It consists of a list of \texttt{RigidTransform}s while a single entry is formed by a \texttt{HomogeneousTransformationMatrix} and an associated \texttt{TimeStamp}. 
The values for the geometric data, the transformation matrix and the time stamps are accompanied with units of measurements.
		\begin{figure}
			\includegraphics[scale=0.84]{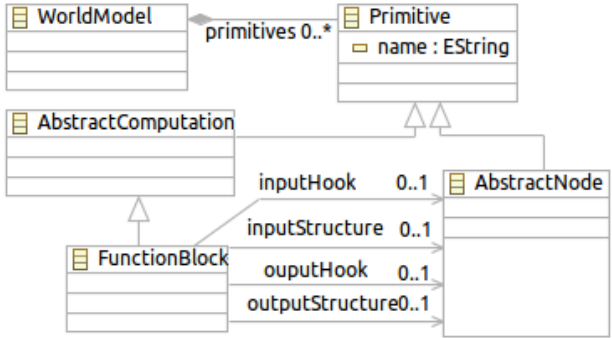}
			\caption{Ecore model of function blocks for data processing. Input and output data is specified via \emph{hooks} and \emph{structure} definitions. Hooks describe where the data is located at run time, while structure definition describes how the data looks like.}
			\label{fig::rsg_ecore_function_block}
		\end{figure}

The \texttt{FunctionBlock} model (cf. Fig. \ref{fig::rsg_ecore_function_block}) to represent the behavioral aspect of the world model consists of four references to \texttt{AbstractNode}s:
An \texttt{inputHook}, an \texttt{inputStructure}, an \texttt{outputHook} and an \texttt{outputStructure}.
The \emph{hooks} reefer to a subgraph at run-time that is to be consumed for further processing or it defines where to add the results of a computation to the scene graph.
The \emph{structure} property represents at design time the expected structure of a scene that is required for an encapsulated algorithm. For example, a function block that implements an algorithm for segmentation of point cloud data can have a \texttt{PointCloud} node as structural input. As output structure it provides \emph{any number} of \texttt{Transform} nodes pointing to the centroids of the segmented point clouds. Each \texttt{Transform} has a \texttt{PointCloud} as child node to represent a single segment. To be able to express such multiplicities in the input and output structures the DSL foresees a \emph{cardinality} attribute that is available in the \texttt{AbstractNodeProperties}. 

Function blocks can be used to create processing chains. All intermediate results of such a chain are stored in the scene graph. The input and output structures allows to check on the M1 level if the output of one function block matches as input for a successor function block.  
A trigger mechanism that can execute function blocks based on changes in the scene or based on signaling by other function blocks is planned as future work.


\subsection{M0: Code generation}

Xtend is used to realize the model to text transformation from the M1 to the M0 level. 
As the world model primitives are available in the RSG implementation the code generation for them is a straight forward mapping to the respective API calls. The RSG-DSL has no assumptions on the order of primitives. On the implementation level, children can not be added to parents that will be created afterwards. To overcome this hurdle the transformation uses a depth-first search based graph traversal for the model primitives to ensure correct order of creation. 

Adding a new primitive with the help of the API will return a unique ID which will be kept in a variable that is labeled with the same name as in the model. These corresponding variables improve readability of the generated code on the one hand and keep the unique ID property on the other hand.

The primitives that are in the subgraph of the \texttt{root} node of the \texttt{WorldModel} will be stored in a \emph{SceneSetup.h} file to represent the application specific scene. This file can be included and used within the application. 

All \texttt{FunctionBlock}s result in dedicated header files for each generated interface. An implementation for a function block has to inherit from such an interface. This strategy is inspired by the \emph{Implementation Gap Pattern} \cite{fowler10domain-specific} and it separates generated code from hand-written code via inheritance.


\section{Examples}
\label{sect::examples}

To illustrate the capabilities of the RSG-DSL a set of examples on the M1 and the M0 level is given below. 

\subsection{A robot application scene}
\label{subsect::example_app_scene}

Listing \ref{lst::app_scene} demonstrates an application scene that consists of a subgraph for a sensor and a kitchen table attached to the \emph{group1} \texttt{Group}. The table could be a part of the environment that is expected to be there but its exact position has to be further deduced by some function block. 
The \texttt{root} keyword defines the application scene subgraph. For the sake of readability the structure for the robot carrying the sensor is omitted and subsumed by a single \emph{worldToCamera} \texttt{Transform}. Note that the transform data is accompanied by units of measurements (cf. lines $20$ to $22$). In case of a moving sensor with respect to the world frame further transform data has to be inserted into the cache. Here the provided information given by the RSG-DSL can be seen as an initial value. The \emph{sensor} \texttt{Group} is supposed to be the place where online sensor data will be hooked in that might serve as input for a function block.

An excerpt of the resulting model to text transformation is presented in Listing \ref{lst::m0_code}. The respective API method invocations for \emph{group1} \texttt{Group} and \emph{worldToCamera} \texttt{Transform} are shown. Lines $2$ to $4$ indicate the mapping of M1 level node names to IDs on the M0 level.   

\begin{lstlisting}[language=Java, style=rsg, caption={Application scene setup represented with the RSG-DSL.}, label=lst::app_scene, firstline=0, lastline=49]
root rootNode // application scene

Group rootNode {
	child group1
	child worldToCamera
}

Group group1 {
	Attribute ("name", "scene_objects")
	child kitchenTable
}

Transform worldToCamera {
	Attribute ("name", "wm_to_sensor_tf")
	child sensor
	transforms {
		RigidTransform t1 {
			stamp TimeStamp ( 0.0 s)
			value HomogeneousTransformationMatrix (
			[1.0, 0.0, 0.0, 0.0 m ],
			[0.0, 1.0, 0.0, 0.0 m ],
			[0.0, 0.0, 1.0, 1.0 m ],
			[0.0, 0.0, 0.0, 0.0   ])												   
		}
	}
}

Group sensor {
	Attribute ("name", "sensor")
}
















\end{lstlisting}

\begin{lstlisting}[language=C++, style=C++, caption={Excerpt from generated code for M0 level. Some comments and additional line breaks have been added after generation.}, label=lst::m0_code, float=*]
std::vector<rsg::Attribute>	attributes; // Instantiation of list of attributes.
unsigned int rootNodeId;                // IDs correspond to names in model on M1 level.
unsigned int group1Id;
unsigned int worldToCameraId;
// [...]

/* Add group1 as a new node to the scene graph */
attributes.clear();
attributes.push_back(Attribute ("name", "scene_objects"));
wm->scene.addGroup(rootNodeId, group1Id, attributes); // group1Id is an output parameter
// [...]                                              // and returns a unique ID.

/* Add worldToCamera as a new node to the scene graph */
attributes.clear();
attributes.push_back(Attribute ("name", "wm_to_sensor_tf"));
brics_3d::IHomogeneousMatrix44::IHomogeneousMatrix44Ptr worldToCameraInitialTf(
	new brics_3d::HomogeneousMatrix44(  // Instantiation of HomogeneousTransformationMatrix primitive.
		1.0, 0.0, 0.0, 
		0.0, 1.0, 0.0, 
		0.0, 0.0, 1.0,
		0.0 * 1.0, 0.0 * 1.0, 1.0 * 1.0  // Values are scaled to SI unit [m]. 
));

wm->scene.addTransformNode(rootNodeId, worldToCameraId, attributes, worldToCameraInitialTf, 
		brics_3d::rsg::TimeStamp(0.0, Units::Second) // Value is scaled to SI unit [s]. 
);



\end{lstlisting}

\subsection{Scene structure for a semantic entity}

\begin{figure}
			\centering
			\includegraphics[scale=0.30]{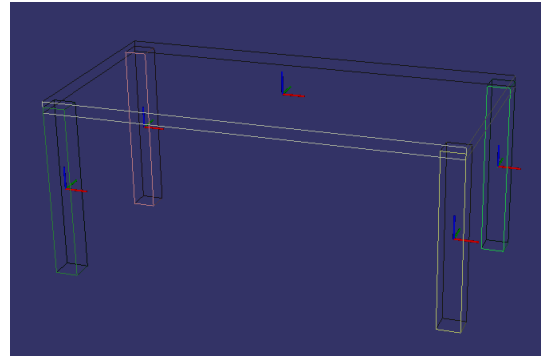}
			\caption{3D visualization of the kitchen table.}
			\label{fig::kitchen_table}			
		\end{figure}

\begin{figure*}[htbp]
			\centering
			\includegraphics[scale=0.28]{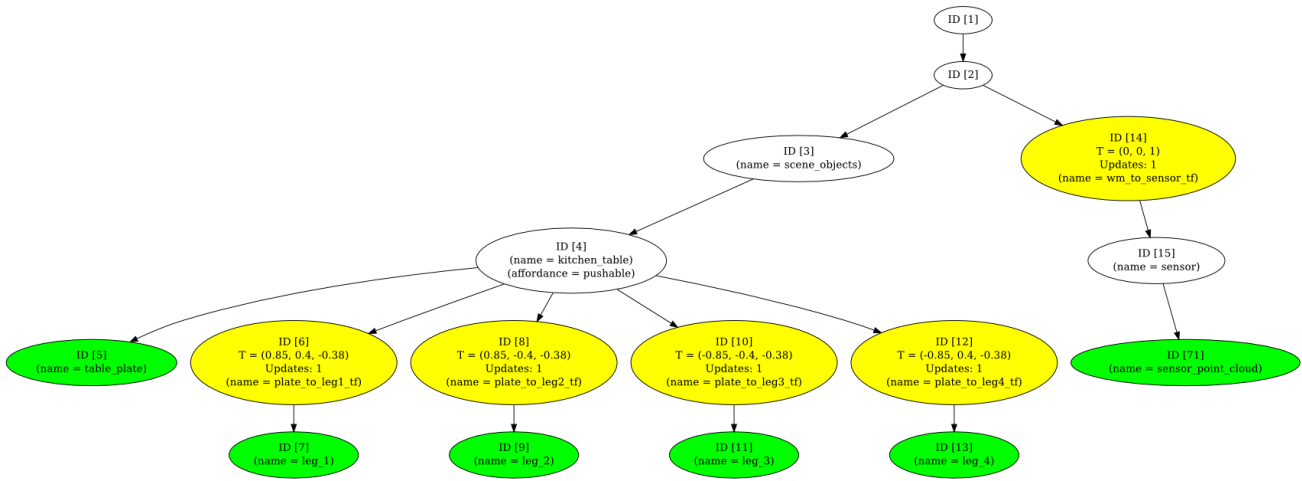}
			\caption{Scene graph structure for the application scene including the kitchen table. Yellow nodes show \texttt{Transform}s while green nodes indicate \texttt{GeometricNode}s. The on M0 level generated IDs are shown in square brackets. Attached attributes are given in brackets. In addition the \texttt{Transform} nodes indicate the translational values T = ($x$, $y$, $z$) and the size of the temporal cache via the \emph{Updates} field.}	
			\label{fig::current_graph}			
		\end{figure*}

As an example for a semantic entity a table is defined in Listing \ref{lst::kitchen_table}. It relates the geometric parts into a scene structure. All legs have a spatial relation from the center of the \emph{tablePlate} defined by the \texttt{Transform} node that is a child of the \emph{kitchenTable} \texttt{Group} node. The example shows only one table leg but the other definitions follow analogously. 
The results are depicted in the Fig. \ref{fig::kitchen_table} and Fig. \ref{fig::current_graph}. The used visualization functionality for the graph structure and the 3D visualization are part of the RSG implementation and demonstrate that the model to text transform of the example works as expected. 

\begin{lstlisting}[language=Java, style=rsg, caption={Kitchen table represented with the RSG-DSL.}, label=lst::kitchen_table, firstline=0, lastline=49]
Group kitchenTable {
	Attribute ("name", "kitchen_table")
	Attribute ("affordance", "pushable")
	child tablePlate	
	child leg1tf 
	child leg2tf
	child leg3tf
	child leg4tf	
}

GeometricNode tablePlate {
	Attribute ("name", "table_plate")
	geometry tablePlateGeometry
}

Box tablePlateGeometry {
	sizeX 1.80 m
	sizeY 0.90 m
	sizeZ 5.0 cm
}

Box tabelLegGeometry {
	sizeX 0.1 m
	sizeY 0.1 m
	sizeZ 0.76 m
}

Transform leg1tf {
	Attribute ("name" , "plate_to_leg1_tf")
	child leg1geom
	transforms {
		RigidTransform t1 {
			stamp TimeStamp ( 0.0 s )
			value HomogeneousTransformationMatrix (
			[1.0, 0.0, 0.0, 0.85 m ] ,
			[0.0, 1.0, 0.0, 0.40 m ] ,
			[0.0, 0.0, 1.0, -0.38 m ] ,
			[0.0, 0.0, 0.0, 0.0   ])
		}	
	}
}

GeometricNode leg1geom {
	Attribute ("name", "leg_1")
	geometry tabelLegGeometry
}

// The other three table legs 
// are set up analogously.
\end{lstlisting}

\subsection{Interface definition for a function block}

A \texttt{FunctionBlock} definition for a point cloud based segmentation algorithms is depicted in Listing \ref{lst::function_block}. The input structure reefers to a point cloud node that contains an internal representation based on the Point Cloud Library (PCL) \cite{rusu113d}. Input and output point clouds are of the same type as shown in lines $7$ and $8$. As output structure a \emph{planes} \texttt{Group} node is specified  
that can have zero or more \texttt{Transform}s that are supposed to point to the centroids of the calculated point cloud segments. Line $21$ reflects this variability by using the optional \texttt{cardinality} keyword. In this case the "*" terminal symbol has the semantics of \emph{any number}.. 
According to the \texttt{outputHook} in line $44$ all results will be inserted to the scene graph as child node of the \emph{sensor} node (cf. Section \ref{subsect::example_app_scene}). An implementation of the function block can be achieved with functionality offered by PCL for instance. Algorithmic details are beyond the scope of this paper. Other point cloud processing libraries could have been chosen as well. Whatever choice the application programmer has been made, it is explicitly represented in the model on the M1 level. \\ 

\begin{lstlisting}[language=Java, style=rsg, caption={A function block represented with the RSG-DSL.}, label=lst::function_block, firstline=0, lastline=49]
PointCloudType PointCloudPCL {
	type "pcl::PointCloud<PointType>"
	sharedPtr "pcl::PointCloud<PointType>::Ptr"
	library "pcl"
}

PointCloud inputCloud type PointCloudPCL
PointCloud planeCloud type PointCloudPCL

GeometricNode pointCloud {
	Attribute ("name", "point_cloud")
	geometry inputCloud
}

Group planes {
	Attribute ("name", "planes")
	child tfToPlaneCentroid
}

Transform tfToPlaneCentroid {
	cardinality *
	child horizontalPlane
	transforms {
		RigidTransform t1 {
			stamp TimeStamp ( 0.0 s)
			value HomogeneousTransformationMatrix (
			[1.0, 0.0, 0.0, 0.0 m ],
			[0.0, 1.0, 0.0, 0.0 m ],
			[0.0, 0.0, 1.0, 0.0 m ],
			[0.0, 0.0, 0.0, 0.0   ])
		}
	}
}

GeometricNode horizontalPlane {
	Attribute ("name", "plane")
	geometry planeCloud
}

FunctionBlock horizontalPlaneSegmentation {
	inputStructure pointCloud 
	inputHook sensorPointCloud
	outputStructure planes
	outputHook sensor	
}
\end{lstlisting}


\section{Conclusion}
\label{sect::conclusion}

This work has presented the RSG-DSL: a DSL for a robotic world model based on the Robot Scene Graph (RSG). It is grounded in executable behavior as code can be generated to be used with an API for an existing implementation of the RSG approach. The RSG-DSL allows to express 
(a) application specific scene setups,
(b) semantic scene structures and
(c) inputs and outputs for the function blocks which are a part of the world model approach.

The RSG-DSL makes a contribution to improve the robot development work flow as world model aspects can be 
explicitly represented in a model-driven tool chain. 
Thus, a developer can create a robotic application quicker and less error prone. 

Future work will include extension of the RSG-DSL approach by multiple levels of detail representations for geometries, uncertainty representations and trigger entities for function blocks. 
Currently the scene setup definition is centered around a single robot system. Language support for distributed and multi-robot applications are  important improvements for the proposed DSL.
The inclusion of other existing DSLs like the geometric relations DSL is a promising research direction with the goal of contributing to a \emph{robotic DSL} that can be composed of a set of languages representing various robotic subfields like world modeling, planning, perception, reasoning or coordination.


\section*{Acknowledgements}

\begin{footnotesize}
The authors acknowledge the fruitful discussions at the 4th International Workshop on Domain-Specific Languages and models for ROBotic systems (DSLRob-13) co-located with IEEE/RSJ IROS 2013, Tokyo, Japan. Insights from the discussions have lead to a clarified version of this paper. 

The authors acknowledge the support from the KU\,Leuven Geconcerteerde
Onderzoeks-Acties \emph{Model based intelligent robot systems} and
\emph{Global real-time optimal control of autonomous robots and mechatronic
systems}, and from the European Union's 7th Framework Programme
(FP7/2007--2013) projects \emph{BRICS} (FP7-231940), \emph{ROSETTA} (FP7-230902), \emph{RoboHow.Cog}
(FP7-288533), and \emph{SHERPA} (FP7-600958). 

\end{footnotesize}


\bibliographystyle{IEEEtran} %
\bibliography{wm_dsl_literature}

\end{document}